\documentclass[10pt]{article} 
\usepackage[preprint]{tmlr}

\usepackage{amsmath,amsfonts,bm}

\def\eqref#1{equation~\ref{#1}}

\def\1{\bm{1}}

\DeclareMathAlphabet{\mathsfit}{\encodingdefault}{\sfdefault}{m}{sl}
\SetMathAlphabet{\mathsfit}{bold}{\encodingdefault}{\sfdefault}{bx}{n}

\usepackage{hyperref}
\usepackage{url}

\usepackage{booktabs}
\usepackage{siunitx}
\usepackage{tabularx}
\usepackage{subfigure}
\usepackage{graphicx}
\usepackage{listings}
\usepackage{svg}

\newcommand*{\ppl}[1]{\num[round-mode=places,round-precision=2]{#1}}
\newcommand*{\usage}[1]{\num[round-mode=places,round-precision=1]{#1}}
\newcommand*{\kl}[1]{\num[round-mode=places,round-precision=2]{#1}}
\newcommand{\topk}{\mathcal{T}_k}
\definecolor{codegreen}{rgb}{0,0.6,0}
\definecolor{codegray}{rgb}{0.5,0.5,0.5}
\definecolor{codepurple}{rgb}{0.58,0,0.82}
\definecolor{backcolour}{rgb}{0.95,0.95,0.92}

\lstdefinestyle{mystyle}{
  backgroundcolor=\color{backcolour}, commentstyle=\color{codegreen},
  keywordstyle=\color{magenta},
  numberstyle=\tiny\color{codegray},
  stringstyle=\color{codepurple},
  basicstyle=\ttfamily\footnotesize,
  breakatwhitespace=false,         
  breaklines=true,                 
  captionpos=b,                    
  keepspaces=true,                 
  numbers=left,                    
  numbersep=5pt,                  
  showspaces=false,                
  showstringspaces=false,
  showtabs=false,                  
  tabsize=2
}

\usepackage{siunitx,etoolbox}
\renewrobustcmd{\bfseries}{\fontseries{b}\selectfont}

\newcommand\equalitytest[3]{%
    \ifnum#1=#2 #3\fi%
}
\newcommand{\memsize}[1]{%
    \equalitytest{#1}{16384}{16k}%
    \equalitytest{#1}{65536}{65k}%
    \equalitytest{#1}{147456}{147k}%
    \equalitytest{#1}{262144}{262k}%
    \equalitytest{#1}{589824}{590k}%
    \equalitytest{#1}{1048576}{1M}%
}
\def \mysp {\hspace{10pt}}

\title{Mixture of A Million Experts}

\author{\name Xu Owen He \email hexu@google.com \\
      \addr Google DeepMind}

\begin{document}

\maketitle

\begin{abstract}
The feedforward (FFW) layers in standard transformer architectures incur a linear increase in computational costs and activation memory as the hidden layer width grows. Sparse mixture-of-experts (MoE) architectures have emerged as a viable approach to address this issue by decoupling model size from computational cost. The recent discovery of the fine-grained MoE scaling law shows that higher granularity leads to better performance. However, existing MoE models are limited to a small number of experts due to computational and optimization challenges. This paper introduces PEER (parameter efficient expert retrieval), a novel layer design that utilizes the product key technique for sparse retrieval from a vast pool of tiny experts (over a million). Experiments on language modeling tasks demonstrate that PEER layers outperform dense FFWs and coarse-grained MoEs in terms of performance-compute trade-off. By enabling efficient utilization of a massive number of experts, PEER unlocks the potential for further scaling of transformer models while maintaining computational efficiency.
\end{abstract}

\begin{figure}[ht]
  \centering
  \subfigure[$6e18$ FLOPs]{\includegraphics[width=0.49\textwidth]{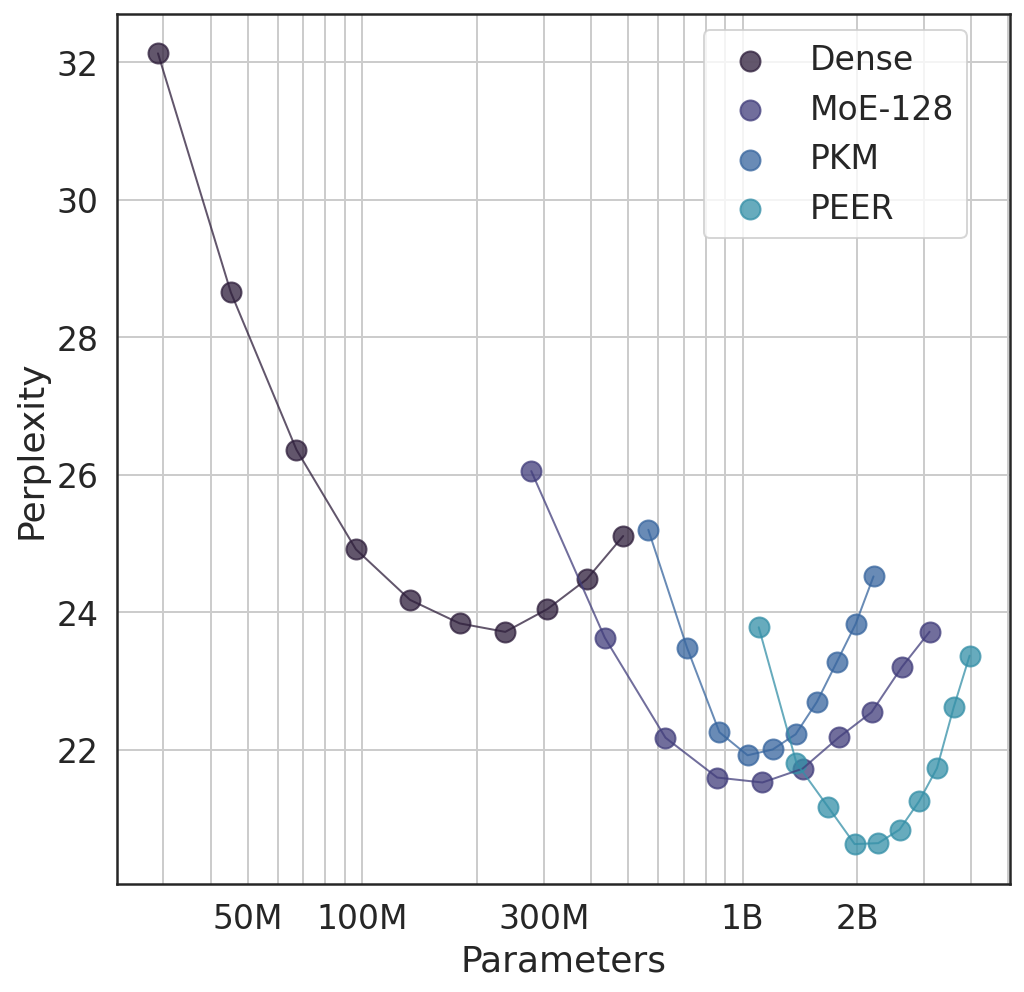}}
  \hfill
  \subfigure[$2e19$ FLOPs]{\includegraphics[width=0.49\textwidth]{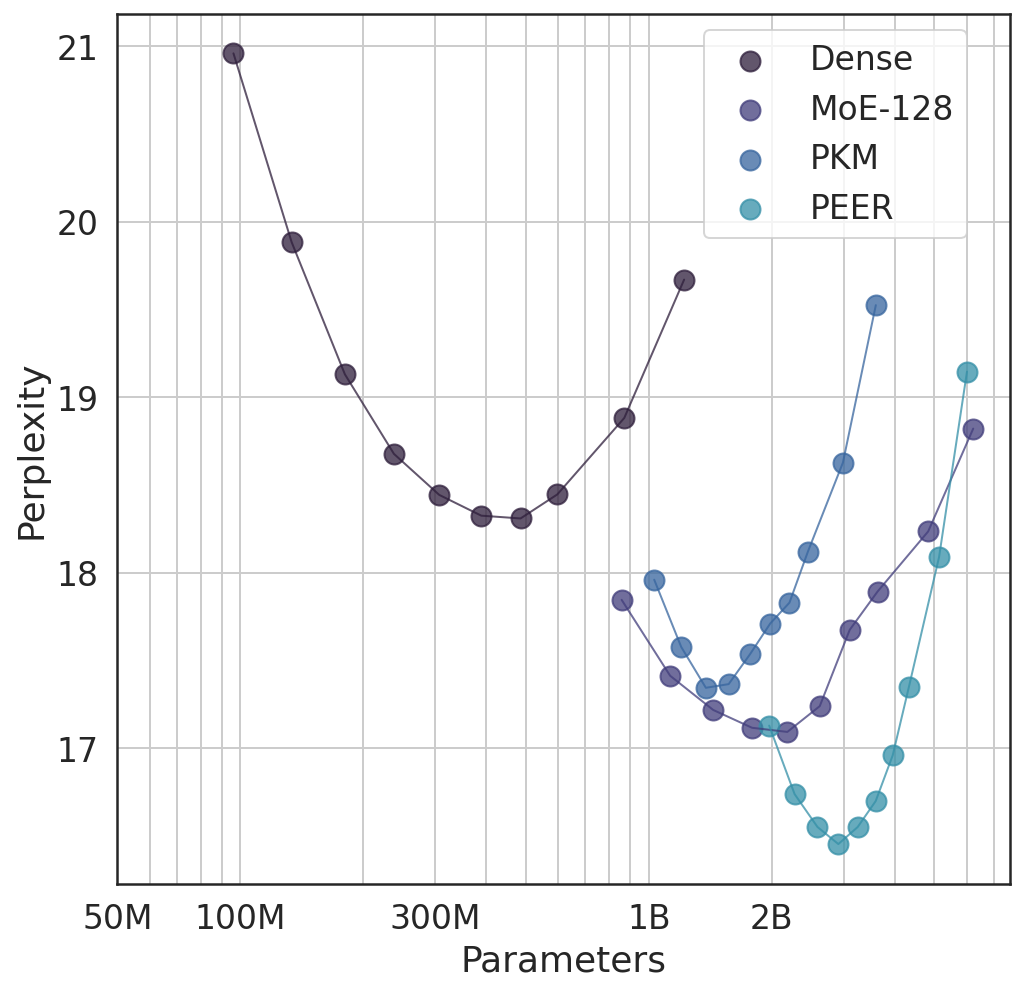}}
  \caption{Isoflop comparison on the C4 dataset between PEER and other baselines with two different FLOP budgets ($6e18$ and $2e19$ FLOPs). The $x$ axis is in log scale.}
  \label{fig:isoflops}
\end{figure}

\section{Introduction}

The past few years have seen the power of scaling \citep{kaplan2020scaling, hoffmann2022training}: increasing the number of parameters, amount of training data, or the computational budget has proven to be a reliable way to improve model performance. Notably, feedforward (FFW) layers, responsible for storing factual knowledge \citep{geva-etal-2021-transformer, dai2022knowledge}, account for two-thirds of the total parameters in a transformer. However, one drawback of these dense FFWs is that their computational footprint (FLOPs and device memory consumption) is linearly proportional to their parameter count. 

To break the coupling between computational cost and parameter count, many recent works \citep{shazeer2017, lepikhin2020gshard, fedus2022switch, zhou2022mixture} have adopted the Mixture-of-Experts (MoE) architecture, which uses a set of sparsely activated expert modules (often FFWs) in place of a single dense FFW. \citet{clark2022unified} studied the scaling law of MoE language models and showed that increasing the number of experts is an effective way to improve performance without increasing the inference cost. However, their experiments showed that the efficiency gains provided by MoEs plateau after a certain model size is reached.  More recently, \citet{krajewski2024scaling} discovered that this plateau was caused by using a fixed number of training tokens. When the number of training tokens is compute-optimal, MoEs consistently outperform dense models in terms of FLOP efficiency. Moreover, they introduced granularity (the number of active experts) as a new scaling axis and empirically showed that using higher granularity improves performance. Extrapolating this fine-grained MoE scaling law suggests that continued improvement of model capacity will ultimately lead to a large model with high granularity, corresponding to an architecture of an immense number of tiny experts.

Beyond efficient scaling, another reason to have a vast number of experts is lifelong learning, where MoE has emerged as a promising approach \citep{aljundi2017expert, chen2023lifelong, Yu_2024_CVPR, li2024theorymixtureofexpertscontinuallearning}. For instance, \citet{chen2023lifelong} showed that, by simply adding new experts and regularizing them properly, MoE models can adapt to continuous data streams. Freezing old experts and updating only new ones prevents catastrophic forgetting and maintains plasticity by design. In lifelong learning settings, the data stream can be indefinitely long or never-ending \citep{mitchell2018never}, necessitating an expanding pool of experts. 

Although both efficient scaling and lifelong learning require MoE designs capable of handling a vast number of experts, to the best of our knowledge, the only architecture supporting more than ten thousands of experts is the Mixture of Word Experts (MoWE) \citep{santos2023memory}. However, MoWE is language-specific and uses a fixed routing scheme. Theoretical and empirical evidence \citep{clark2022unified, dikkala2023on} highlights the advantages of learned routers over non-trainable ones. Thus, an MoE design with a learned router scalable to over a million experts remains an open area for exploration.

This work introduces the Parameter Efficient Expert Retrieval (PEER) architecture, leveraging product key retrieval \citep{lample2019large} for efficient routing to an extremely large number of experts, decoupling computational cost from parameter count. This design demonstrates a superior compute-performance trade-off in our experiments, positioning it as a competitive alternative to dense FFW layers for scaling foundation models. The main contributions of this work are:
\begin{itemize}
    \item \textbf{Exploration of Extreme MoE Setting:} Deviating from the focus on a small number of large experts in previous MoE research, this work investigates the under-explored case of numerous tiny experts.
    \item \textbf{Learned Index Structure for Routing:} Demonstrating for the first time that a learned index structure \citep{kraska2018case} can efficiently route to over a million experts.
    \item \textbf{New Layer Design:} Combining product key routing with single-neuron experts, we introduce the PEER layer that expands layer capacity without significant computational overheads. Empirical results demonstrate its superior efficiency compared to dense FFW, coarse-grained MoEs and Product Key Memory (PKM) layers.
    \item \textbf{Comprehensive Ablation Studies:} We investigate the impact of different design choices of PEER such as number of experts, active parameters, number of heads and query batch normalization on language modeling tasks.
\end{itemize}

\begin{figure}[ht]
    \centering
    \includegraphics[width=0.95\linewidth]{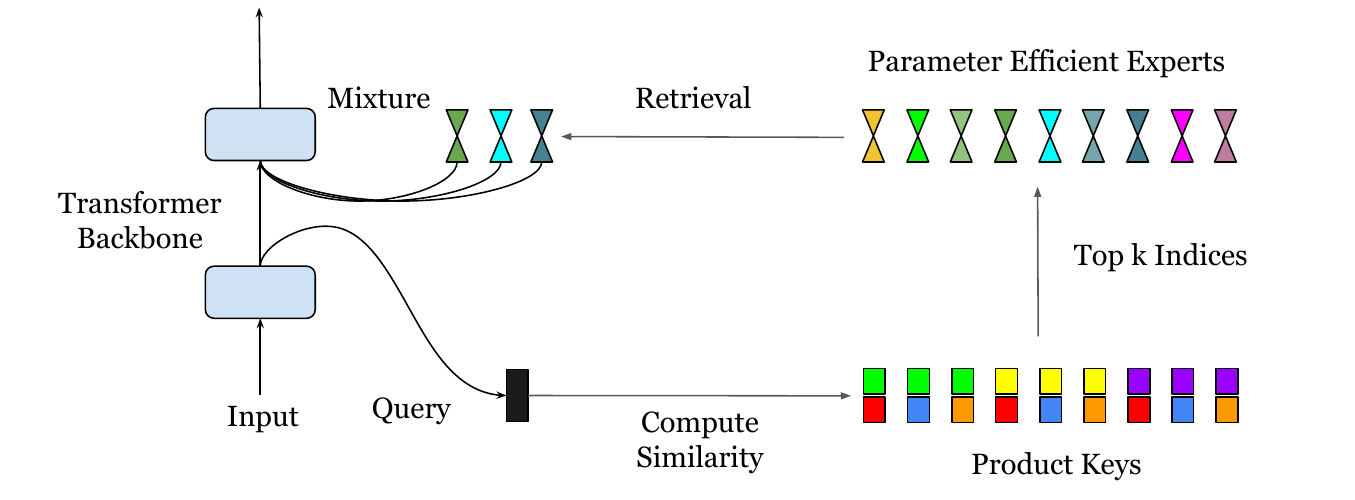}
    \caption{\textbf{Illustration of the PEER layer.} A PEER layer can be inserted in the middle of a transformer backbone or can be used to replace FFW layers. Given the state vector $x$ from the previous layer, a query network $q$ maps it to a query vector $q(x)$, which is then compared with the product keys to compute the router scores and to retrieve the top $k$ experts $e_1, ..., e_k$. After the retrieved experts make their predictions $e_i(x)$, their outputs are linearly combined using the softmax-normalized router scores as weights.}
    \label{fig:peer_layer}
    \vspace{-0.0cm}
\end{figure}

\section{Method}
In this section, we introduce the Parameter Efficient Expert Retrieval (PEER) layer, which is a Mixture of Experts architecture using product keys \citep{lample2019large} in the router and single-neuron MLPs as experts. Fig. \ref{fig:peer_layer} illustrates the computational process within a PEER layer.

\paragraph{PEER Overview} Formally, a PEER layer is a function $f:\mathbb{R}^n\to\mathbb{R}^m$ that consists of three parts: a pool of $N$ experts $\mathbb{E}:=\{e_i\}_{i=1}^{N}$, where each expert $e_i:\mathbb{R}^n\to\mathbb{R}^m$ shares the same signature as $f$, a corresponding set of $N$ product keys $\mathbb{K}:=\{k_i\}_{i=1}^{N}\subset\mathbb{R}^d$, and a query network $q:\mathbb{R}^n\to\mathbb{R}^d$ that maps the input vector $x\in\mathbb{R}^n$ to a query vector $q(x)$.
Let $\topk$ denote the top-k operator. Given an input $x$, we first retrieve a subset of $k$ experts whose corresponding product keys have the highest inner products with the query $q(x)$.
\begin{align}
    \mathbb{I} &= \topk \left( \{q(x)^T k_i\}_{i=1}^{N} \right) \label{eq:topk} &\text{\# Retrieve top $k$ experts~\quad}
\end{align}
Then we apply nonlinear activations (such as softmax or sigmoid)  to the query-key inner products of these top $k$ experts to obtain the router scores. 
\begin{align}
    g_i(x)&= s(q(x)^Tk_i)\label{eq:router} &\text{\# Compute router scores~\quad \ }
\end{align}
Finally, we compute the output by linearly combining the expert outputs weighted by the router scores.
\begin{align}
    f(x) &= \sum\nolimits_{i \in \mathbb{I}} g_i(x) e_i(x) \label{eq:moe} &\text{\# Aggregate expert outputs \ }
\end{align}
\paragraph{Product Key Retrieval}
Since we intend to use a very large number of experts ($N\geq 10^6$), naively computing the top $k$ indices in Eq. \ref{eq:topk} can be very expensive. Hence we apply the product key retrieval technique here. Instead of using $N$ independent $d$-dimensional vectors as our keys $k_i$, we create them by concatenating vectors from two independent sets of $\frac{d}{2}$-dimensional sub-keys $\mathbb{C},\mathbb{C}'\subset \mathbb{R}^{\frac{d}{2}}$:
\begin{equation}
    \mathbb{K}=\{\begin{bmatrix} c \\ c' \end{bmatrix} |c\in\mathbb{C}, c'\in\mathbb{C}'\}
\end{equation}
Note that here $\mathbb{C},\mathbb{C}'$ have cardinality $\sqrt{N}$ and $c, c'$ have dimensionality  $\frac{d}{2}$. So in practice, we choose $N$ to be a perfect square and $d$ to be an even number. 

This Cartesian product structure of $\mathbb{K}$ allows us to find the top $k$ experts efficiently. Instead of comparing $q(x)$ to all $N$ keys in $\mathbb{K}$ and selecting the top k matches, we can split the query vector $q(x)$ into two sub-queries $q_1$ and $q_2$ and apply the top k operations to the inner products between the sub-queries and sub-keys respectively:
\begin{equation}
    \mathbb{I}_\mathbb{C} = \topk \left( (q_1^T c_i) \right)
    ,\quad \quad \mathbb{I}_{\mathbb{C}'} = \topk \left( (q_2^T c_j') \right) \\
\end{equation}
This results in a set of $k^2$ candidate keys $\mathbb{K}':=\{ \begin{bmatrix} c_i \\ c_j \end{bmatrix}  |i\in\mathbb{I}_\mathbb{C}, j\in\mathbb{I}_\mathbb{C}'\}$, and it is mathematically guaranteed that the $k$ most similar keys to $q(x)$ from $\mathbb{K}$ are in this candidate set. Moreover, the inner product between the candidate key and $q(x)$ is simply the sum of inner products between the sub-keys and sub-queries: $q(x)^T\begin{bmatrix} c_i \\ c_j \end{bmatrix} = q_1^Tc_i+q_2^Tc_j$. Hence we can apply the top-k operator again to these $k^2$ inner products to get the top k matching keys from the original set of product keys $\mathbb{K}$. As explained in \cite{lample2019large}. This reduces the complexity of top k expert retrieval in Eq. \ref{eq:topk} from $O(Nd)$ as done naively by exhaustive search to $O((\sqrt{N}+k^2)d)$.

\paragraph{Parameter Efficient Experts and Multi-Head Retrieval}
 Unlike other MoE architectures, which often set the hidden layer of each expert to the same size as other FFW layers, in PEER, every expert $e_i$ is a singleton MLP, in other words, it has only one hidden layer with a single neuron:
 \begin{equation}
     e_i(x):=\sigma(u_i^T x)v_i
 \end{equation}
where $v_i, u_i$ are not matrices but vectors with the same dimension as $x$, and $\sigma$ is a nonlinear activation function such as ReLU or GELU. We omit bias terms here for brevity.

Instead of varying the size of individual experts, we adjust the expressiveness of a PEER layer by using multi-head retrieval, similar to the multi-head attention mechanism in transformers and the multi-head memory in PKMs. In particular, we use $h$ independent query networks instead of one, each computes its own query and retrieves a separate set of $k$ experts. However, different heads share the same pool of experts with the same set of product keys. The outputs of these $h$ heads are simply summed up:
\begin{equation}
    f(x) := \sum_{i=1}^hf^i(x) =  \sum_{i=1}^h\sum_{j\in\mathbb{I}^i}g_j(x)e_j(x)
\end{equation}
One can verify that when only one expert is retrieved ($k=1$) per head, using a PEER layer with $h$ heads is the same as using one expert with $h$ hidden neurons:
\begin{equation}
    f(x) = \sum_{i=1}^he^i(x) =  \sum_{i=1}^h\sigma(u_i^T x)v_i= V\sigma(W^Tx);
\end{equation}
where $W=[u_1, \cdots, u_h], V=[v_1, \cdots, v_h]$. In other words, PEER dynamically assembles an MLP with $h$ neurons by aggregating $h$ singleton MLPs retrieved from a shared repository. Compared to existing MoE approaches that use MLPs with multiple hidden neurons as experts, this design allows shared hidden neurons among experts, enhancing knowledge transfer and parameter efficiency.

Algorithm \ref{alg:peer} shows a simplified implementation of the PEER forward pass, storing parameter-efficient expert weights in embedding layers and combining them with einsum operations. This implementation can be easily extended to experts of the GLU variants \citep{shazeer2020gluvariantsimprovetransformer} by adding additional linear gating weights. In practice, an efficient implementation may require  specialized hardware kernels to accelerate embedding lookup and fusion with the einsum operations.

\paragraph{Why A Large Number of Small Experts?}
Given an MoE layer, we can characterize it by three hyperparameters: the total number of parameters $P$, the number of active parameters per token $P_\text{active}$ and the size of a single expert $P_\text{expert}$. \citet{krajewski2024scaling} showed that the scaling law of MoE models has the following form:
\begin{equation}
    \mathcal{L}(P, D, G) = c + (\frac{g}{G^\gamma}+a)\frac{1}{P^\alpha}+\frac{b}{D^\beta},
    \label{equ:scalinglaw}
\end{equation}
where $\mathcal{L}$ is the final test loss, $a, b, g, \gamma, \alpha, \beta$ are constants, $D$ is the total number of training tokens and the granularity $G$ is the number of active experts:
\begin{equation}
    G:=\frac{P_\text{active}}{P_\text{expert}}
\end{equation}
In order to improve model performance, we need to scale up $P, D, G$. On the other hand, it is essential to limit $P_\text{active}$ because the computational and memory costs are primarily determined by the active parameters during training and inference. Notably, the memory footprint corresponding to $P_\text{active}$ has to be multiplied by the number of tokens in a batch, while the memory cost of $P$ is independent of the batch size and sequence length because only one copy of the model needs to be stored. 

As a result, we want to increase $P, G$ but not $P_\text{active}$. Since the expert size $P_\text{expert}=P_\text{active}/G$ and the number of experts $N=P/P_\text{expert}=P\cdot G/P_\text{active}$, this implies that we should decrease the size of each expert, $P_\text{expert}$, and increase the number of experts $N$. Hence we need a large number of small experts.

In general, for experts that are MLPs with a single hidden layer. $P_\text{expert}=(2d_\text{model}+1)d_\text{expert}$ and $P_\text{active}=(2d_\text{model}+1)d_\text{active}$, where $d_\text{model}$, $d_\text{expert}$ and $d_\text{active}$ are the hidden dimension of the transformer, the number of hidden neurons used in one expert and the total number of hidden neurons activated per token, respectively.

In the case of PEER, we use the smallest expert size possible by setting $d_\text{expert}=1$, and the number of activated neurons is the number of retrieval heads multiplied by the number of experts retrieved per head: $d_\text{active}=hk$. Consequently, the granularity of PEER is always $G=P_\text{active}/P_\text{expert}=d_\text{active}/d_\text{expert}=hk$.

\begin{lstlisting}[language=Python, caption={Pseudo code implementation of a PEER layer forward pass.  An example implementation of the get\_indices and query\_proj functions in Pytorch can be found in \citet{pkm_colab}}, label=alg:peer, escapechar=|]
def peer_forward(self, x):
    # Embedding layers storing the down/up projection weights of all experts
    self.w_down_embed = nn.Embed(num_embeddings=self.n_experts, features=self.d_model)
    self.w_up_embed = nn.Embed(num_embeddings=self.n_experts, features=self.d_model)
    
    # Retrieve the weights of the top matching experts using product keys
    # indices and scores have the shape 'bthk', where h is the number of heads
    indices, scores = self.get_indices(self.query_proj(x), self.sub_keys, top_k=self.k)
    w_down = self.w_down_embed(indices)
    w_up = self.w_up_embed(indices)
    
    # Compute weighted average of expert outputs
    x = jnp.einsum('btd, bthkd->bthk', x, w_down)
    x = self.activation(x)
    x = x * nn.softmax(scores)
    x = jnp.einsum('bthk, bthkd->btd', x, w_up)
    return x
\end{lstlisting}

\section{Experiments}
\label{sec:experiment}

\subsection{Pretraining isoFLOP Analysis}
We compare PEER with various baselines using isoFLOP analysis \citep{borgeaud2022improving}. We chose a fixed FLOP budget ($6e18$ and $2e19$) and jointly varied the model size and the number of training tokens from the C4 dataset \citep{raffel2020exploring} to obtain isoFLOP curves. Each point on an isoFLOP curve has the same computational cost, and we plot them in terms of their model size and final validation perplexity on C4.

For the dense baselines, we varied their size by changing the number of layers, attention heads and model dimensions. For MoE, PKM and PEER methods, we took each of the dense models considered and replaced the FFW layer in the middle block (e.g. in a 12 block transformer, we replace the FFN in block 6) by a layer of MoE, PKM and PEER, respectively. 

In MoE, we used the expert-choice \citep{zhou2022mixture} routing algorithm, which effectively addresses the expert load imbalance issue and generally outperforms token-choice MoEs (see Section \ref{sec:related_works} for a review and comparison of these approaches). Each expert has the same size as the original MLPs in the corresponding dense model, and we use $128$ experts to cover the same range of model sizes as our PEER models. This type of MoE represents standard coarse-grained MoE approaches, which consist of a small number of large experts. 

In PKM, we used $1024^2$ memories with $h=8$ heads and top $k=32$ memories were selected per head. We also applied query batch normalization, as recommended in the original PKM paper \citep{lample2019large}, to enhance memory usage.

In PEER, we used $1024^2$ experts with $h=8$ heads and top $k=16$ experts per head. By default, we also enabled query BatchNorm to increase expert usage. Ablation studies in subsection \ref{subsec:ablation} investigate the effect of these hyperparameters. Unlike the expert-choice MoE baseline, PEER represents a fine-grained approach where a large number of small experts are employed.

Across all model sizes and methods, we maintained a consistent batch size (128) and sequence length (2048). We calculated the number of training steps by dividing the total compute budget by the FLOPs per training step. Fig. \ref{fig:isoflops} presents the isoFLOP profiles. Compared to the dense FFW baseline, the sparse alternatives shift the isoFLOP curves downward and to right because they introduce a larger number of total parameters $P$ but utilize a smaller or equal number of active parameters $P_\text{active}$. Given the same compute budget, a PEER model achieves the lowest compute-optimal perplexity.

\subsection{Evaluation on Language Modeling Datasets}
After determining the compute-optimal model for each method based on the isoFLOP curves, we evaluated the performance of these pretrained models on several popular language modeling datasets, including Curation Corpus \citep{curationcorpusbase:2020}, Lambada \citep{paperno-etal-2016-lambada}, the Pile \citep{pile}, Wikitext \citep{merity2016pointer} and the pretraining dataset C4. Table  \ref{tab:other_lm_evals} presents a summary of the evaluation results. We grouped the models based on their FLOP budgets used during training.

\begin{table*}[ht]{
\caption{Perplexities of the compute-optimal models of each method on language modeling datasets.
\label{tab:other_lm_evals}}
} \smallskip
\centering
{\small

\begin{tabular}[b]{l c c c c c}
    \toprule
    Method & Curation  & Lambada & Pile &  Wikitext & C4 \\
           & Corpus    &         &      &           &   \\
    \midrule
    Dense (6e18) & 23.26 & 21.95 & 24.55 & 29.14 & 23.84 \\
    MoE (6e18)    & 20.98 & 19.09 & 23.26 & 26.10 & 21.41 \\
    PKM (6e18)    & 21.80 & 19.39 & 20.49 & 27.09 & 21.92 \\
    PEER (6e18)   & \bfseries 20.68 & \bfseries 17.65 & \bfseries 19.01 & \bfseries  25.48 & \bfseries 20.63 \\  
    \midrule
    Dense (2e19)  & 17.70 & 12.28 & 18.19 & 21.21 & 18.31 \\
    MoE (2e19)    & 16.88 & 12.97 & 17.41 & 20.28 & 17.12 \\
    PKM (2e19)    & 17.03 & 11.18 & 16.34 & 20.26 & 17.36 \\
    PEER (2e19)   & \bfseries 16.34 & \bfseries 10.33 & \bfseries 14.99 & \bfseries 19.09 & \bfseries 16.45 \\
    \bottomrule
\end{tabular}}
\end{table*}

\subsection{Ablations}
\label{subsec:ablation}
\begin{figure}[ht]
  \centering
  \subfigure[Varying Total Expert Num]{\includegraphics[width=0.48\textwidth]{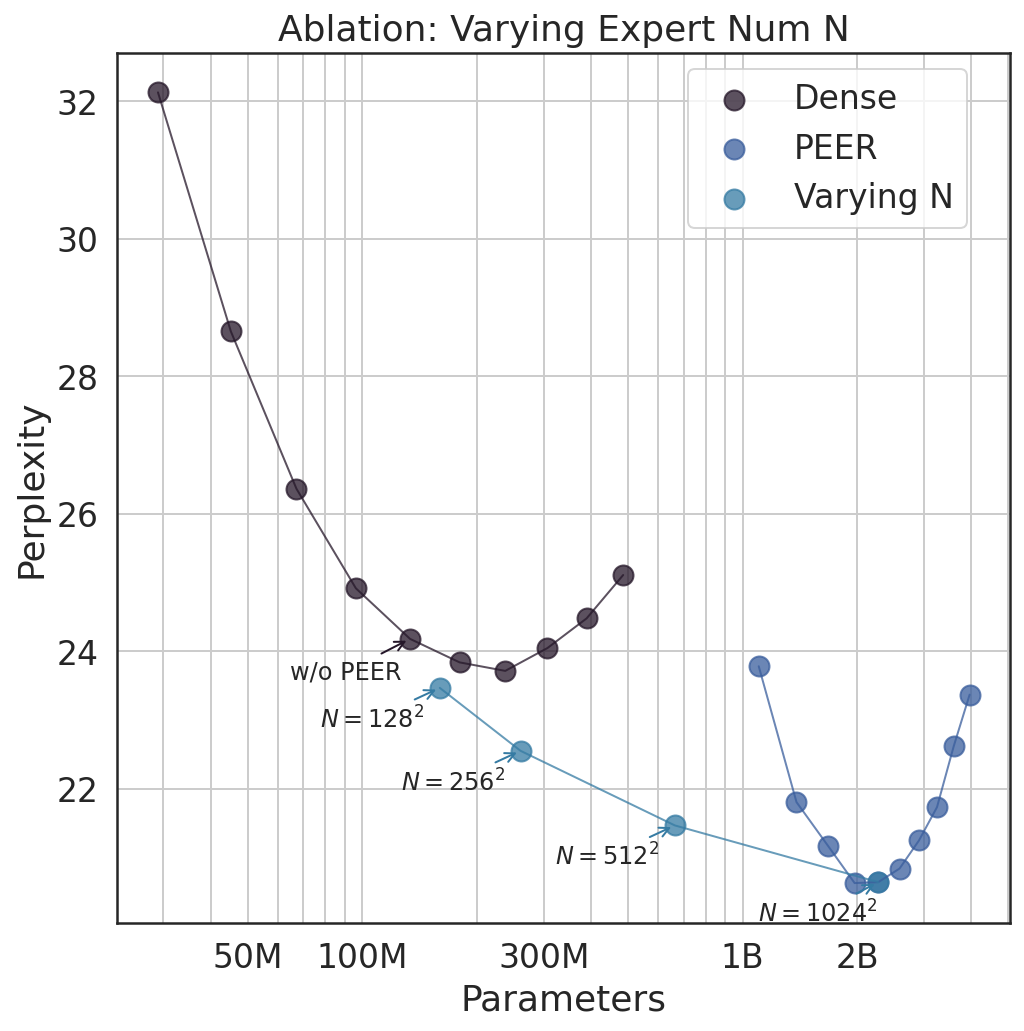}}
  \hfill
  \subfigure[Varying Active Expert Num]{\includegraphics[width=0.5\textwidth]{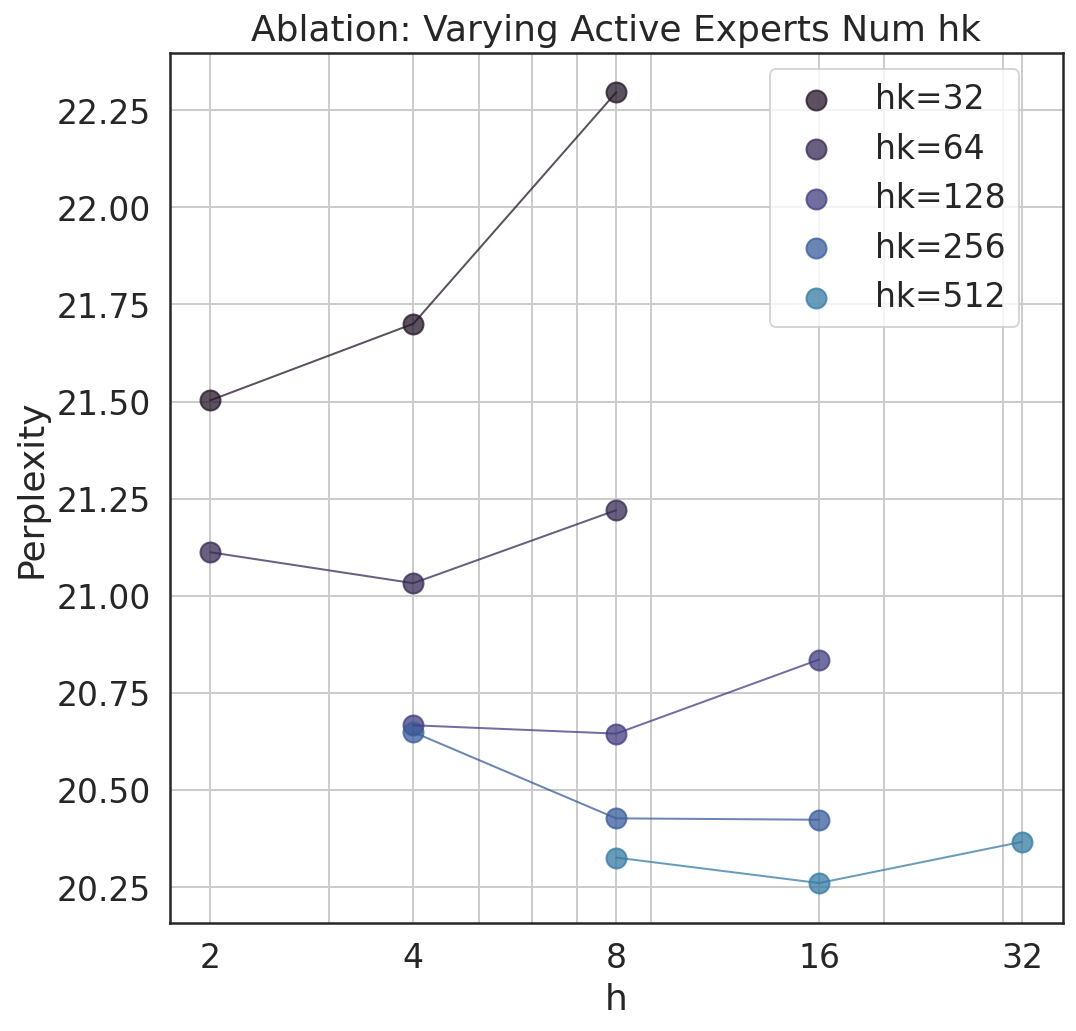}}
  \caption{We conduct two ablation studies using the same PEER model configuration. In (a), we vary the total number of experts $N$ while keeping the same number of active experts $hk=128$. In (b), we vary the number of active experts $G=hk$ by jointly changing $h$ and $k$ while keeping the total number of experts at $N=1024^2$.}
  \label{fig:ablation}
\end{figure}

\paragraph{Varying the Number of Total Experts}
The models in the isoFLOP plot depicted in Fig.~\ref{fig:isoflops} all have over a million ($1024^2$) experts. Here we conduct an ablation study on the effect of the number of experts $N$, which determines the total parameter count $P$ in Eq. \ref{equ:scalinglaw}. We selected the model at the isoFLOP-optimal position and vary the number of experts ($N=128^2, 256^2, 512^2, 1024^2$) in the PEER layer while keeping the number of active experts constant ($h=8, k=16$). The results are shown in Fig. \ref{fig:ablation} (a). As can be seen, the isoFLOP curve interpolates between the PEER model with $1024^2$ experts and the corresponding dense backbone without replacing the FFW layer in the middle block by a PEER layer. This demonstrates that simply increasing the number experts can improve model performance.

\paragraph{Varying the Number of Active Experts}
We also conducted an ablation study on the effect of the number of active experts $hk$, which equals the granularity $G$ in Eq. \ref{equ:scalinglaw}. We systematically varied the number of active experts ($hk=32, 64, 128, 256, 512$)  while keeping the number of total experts constant ($N=1024^2$). Furthermore, for a given $hk$, we jointly varied $h$ and $k$ to identify the optimal composition. The resulting isoFLOP curves, plotted over the number of heads ($h$), are shown in Fig. \ref{fig:ablation} (b).

The results indicate that, within the range of values considered, higher $hk$ generally leads to improved performance. Notably, the optimal $h$ increases as $hk$ increases. However, the performance gradually saturates, and increasing the number of active experts also increases device memory consumption and may necessitate additional accelerator devices. Thus in practice, the appropriate $hk$ values should be selected based on the trade-off between performance, device number and computational resource requirements.

\begin{table*}[ht]{
\caption{\textbf{KL and expert usage for different memory sizes, with and without query BN.} Similar to the findings in PKM, using query BN results in a more balanced usage of the experts.
\label{tab:ablation_expe_num_query_bn}}
} \smallskip
\centering
{\small

\begin{tabular}[b]{l|c@{\mysp}c|c@{\mysp}c|c@{\mysp}c|c@{\mysp}c}
    \toprule
    Expert num $N$ & \multicolumn{2}{c|}{\memsize{16384}} & \multicolumn{2}{c|}{\memsize{65536}} & \multicolumn{2}{c|}{\memsize{262144}} &  \multicolumn{2}{c}{\memsize{1048576}} \\
    BatchNorm   & No              & Yes             & No              & Yes             & No              & Yes             & No              & Yes                      \\
    \midrule
    Perplexity  & \ppl{23.4658}   & \ppl{23.4690}   & \ppl{22.6121}   & \ppl{22.5543}   & \ppl{21.5409}   & \ppl{21.4696}   & \ppl{20.7307}   & \ppl{20.6449}   \\
    Expert Usage (\%)  &   \usage{100}     &   \usage{100}     & \usage{100} &  \usage{100} & \usage{99.9996} & \usage{100} & \usage{99.7930} & \usage{99.9754}  \\
    Unevenness ($\downarrow$)        & \kl{0.4489}     &  \kl{0.2954}     & \kl{0.6325}     & \kl{0.4420}     & \kl{0.9707}     & \kl{0.6552}     & \kl{1.5244}     & \kl{1.0588}        \\
    \bottomrule
\end{tabular}}
\end{table*}

\paragraph{Expert Usage and Query Batch Normalization}
Given the presence of over a million experts in the PEER layer, it is natural to inquire how many of these experts are actually selected during inference and whether their usage is evenly distributed. To analyze this, we kept an accumulated router score, denoted as $z'_i = \sum_x g_i(x)$ for each expert $e_i$ across all tokens $x$ within the C4 validation set. Here $g_i(x)$ is the router score used to aggregate the expert output when token $x$ is given as input, with $g_i(x)=0$ if expert $e_i$ is not selected. From these accumulated router scores, we can obtain an empirical probability distribution vector, denoted as $z=z'/||z'||_1$, representing the distribution of all experts over the C4 validation set. Then we computed the following metrics proposed by \citet{lample2019large} to assess the usage and distribution of experts:
\begin{itemize}
    \item \textit{Expert Usage}: the fraction of experts retrieved during inference: $\#\{z_i\neq 0\}$
    \item \textit{Unevenness}: KL divergence between $z$ and the uniform distribution: $\log(N)+\sum_i z_i\log(z_i)$
\end{itemize}
where $N$ is the number of total experts.

By default, we also added a batch normalization (BN) layer on top of the query network, as proposed by \citet{lample2019large} to increase the expert usage during training. Here we study the effect of adding this BN layer on the above-mentioned metrics. 

Table \ref{tab:ablation_expe_num_query_bn} presents the expert usage and unevenness for varying numbers of experts, with and without BN. We can see that even for 1M experts, the expert usage is close to $100\%$, and using BN can lead to more balanced utilization of the experts and lower perplexities. These findings demonstrate the effectiveness of the PEER model in utilizing a large number of experts.

\begin{figure}[ht]
    \centering
    \includegraphics[width=0.6\linewidth]{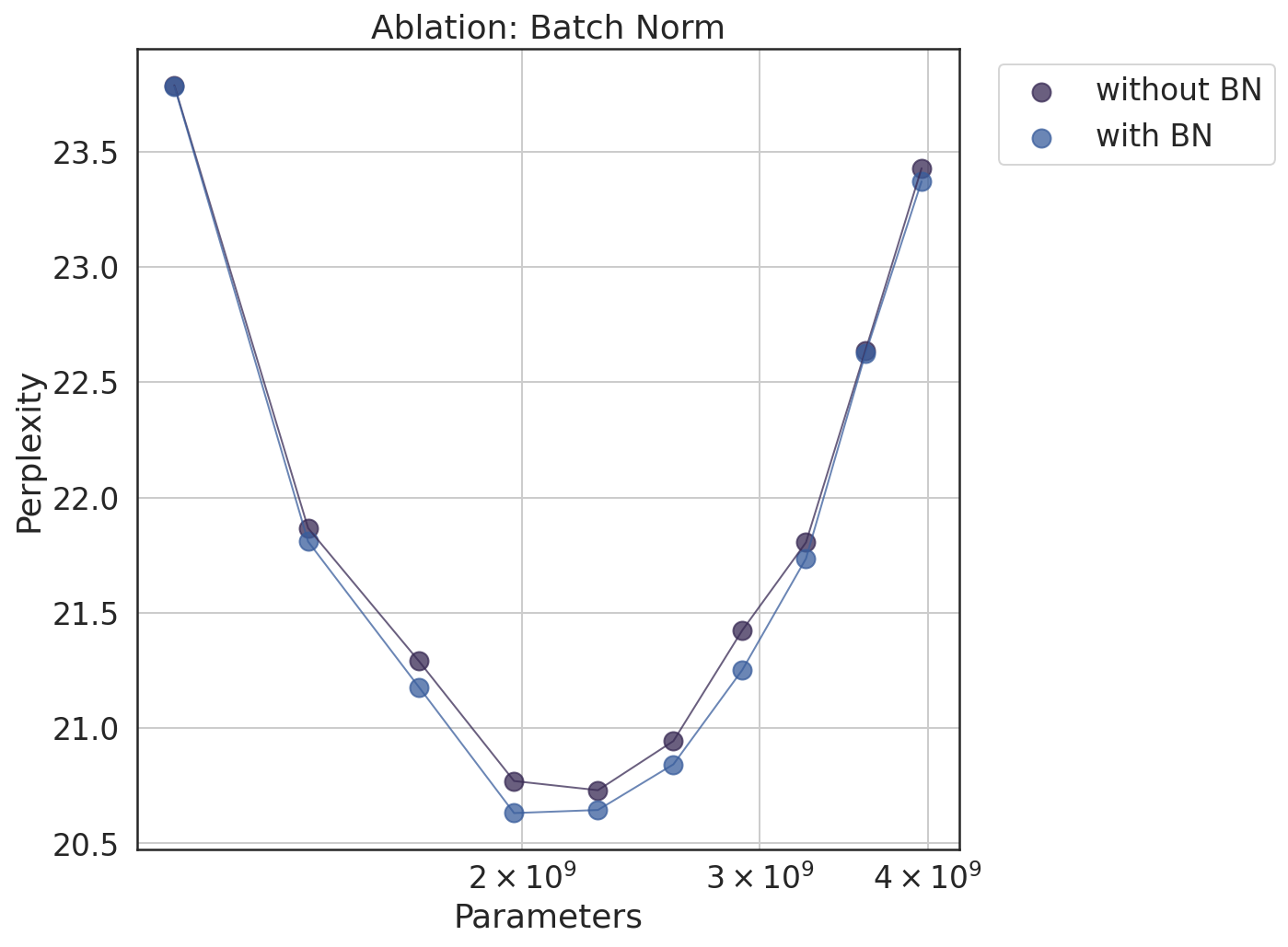}
    \caption{\textbf{Query BatchNorm Ablation.} IsoFLOP curves of a PEER model with 1M experts on the C4 dataset, with and without query BatchNorm.}
    \label{fig:ablationBN}
    \vspace{-0.0cm}
\end{figure}

We additionally compared isoFLOP curves with and without BN. Fig. \ref{fig:ablationBN} shows that the PEER model with BN generally achieves lower perplexities. While the difference is not significant, it is most pronounced around the isoFLOP-optimal region.

\section{Related Works}
\label{sec:related_works}
\paragraph{Mixture of Expert}
Since \citet{shazeer2017} demonstrated the effectiveness of sparsely-gated Mixtures of Experts (MoEs) in efficiently increasing model capacity on GPU clusters, MoEs have emerged as a popular technique for scaling large models efficiently. Subsequent research \citep{fedus2022switch, lepikhin2020gshard, pmlr-v162-du22c} has proposed variations to address challenges such as load balancing, communication overhead, and training instability. These methods usually replace feedforward (FFW) layers in certain Transformer blocks with sparsely-gated MoE layers, which consist of multiple FFW layers as experts. Typically each expert matches the size of the regular dense FFW layer. Gating scores are calculated for each expert and token, and only the top k experts are activated for each token. These methods are known as token-choice methods. More recently, \citet{zhou2022mixture} introduced the Expert Choice routing method, where experts choose the top k tokens instead of tokens selecting experts. However, both token-choice and expert-choice methods require the top-k operator on a gating score matrix of size $N\times M$ ($N$: number of experts, $M$: number of tokens), resulting in a routing cost of at least $O(N)$. This limits their practical application to a small number of experts (typically less than 128). 

Instead of using the top-k operator, some works also proposed using deterministic hash tables as routers \citep{roller2021hash, santos2023memory}. With $O(1)$ average lookup complexity, these methods offer potential scalability to a large number of experts. However, these routers are fixed and not learned. \citet{clark2022unified} showed that deterministic routing does not scale as well as trainable routers. Furthermore, \citet{dikkala2023on} proved theoretically that learned routers offer non-trivial advantages  over their fixed counterparts, such as removing spurious directions and identifying latent clusters in data. In contrast to previous works, the proposed PEER layer employs a learned router with sublinear ($O(\sqrt{N})$) complexity. 
 
Since PEER uses lightweight experts, our work is also related to recent studies on parameter-efficient MoEs \citep{wang-etal-2022-adamix, zadouri2024pushing}. These methods utilize parameter efficient fine-tuning (PEFT) adapters as experts instead of full-sized FFWs. Their focus is on minimizing the number of parameters updated during fine-tuning, allowing storage of only one copy of the large backbone model. In PEER, parameter efficiency refers to the small number of active parameters in the MoE layer, which directly affects FLOPs and activation memory consumption during pre-training and inference. However, PEER could potentially be adapted to retrieve a large number of PEFT adapters.
\paragraph{Retrieval-Augmented Models}
Our proposed method, with its retrieval mechanism for a large number of experts, aligns with the emerging field of retrieval-augmented models. These models facilitate large model memorization by retrieving knowledge from external databases, leading to improved accuracy and efficiency on knowledge-intensive tasks. Some notable works in this domain include ones by \citet{khandelwal2019generalization, pmlr-v162-borgeaud22a, guu2020retrieval}. While these methods retrieve data in various formats, for instance, tokens \citep{khandelwal2019generalization}, chunks \citep{borgeaud2022improving} or knowledge graphs \citep{kang2023knowledge} (see \citep{gao2023retrieval} for a comprehensive survey on this topic), they differ from the proposed method in that they retrieve data rather than learned functions (experts). This distinction sets our parameter-efficient expert retrieval approach apart from existing retrieval-augmented models.

\paragraph{Efficient Feedforward Layers}
Enhancing the efficiency of feedforward networks has been a long-standing area of research. Similar to PEER, most approaches are based on the idea of conditional computation \citep{bengio2013deep}, where a gating mechanism is trained to determine which subset of neurons to compute. For instance, \citet{davis2013low} utilized low-rank weight matrix approximation to estimate the sign of pre-nonlinearity activations. Neurons with negative activations are omitted as they will produce zeros after the nonlinearity. \citet{bengio2015conditional} explored reinforcement learning to develop an activation-dependant policy for dropping blocks of neurons. More recently, \citet{belcak2023fast} introduced the Fast FeedForward (FFF) layer that employs a differentiable balanced binary tree to select a neuron block for computation. During inference, only one leaf (corresponding to one block) is selected, hence it has $O(\log(N))$ complexity, where $N$ is the total number of blocks in the tree. However, during training, all leaves and intermediate nodes are activated for gradient calculation, imposing a training complexity of $O(N)$ and limiting the total number of blocks. The most relevant work to ours is the Product Key Memory (PKM) \citep{lample2019large}, whose retrieval technique is utilized as the router in the PEER layer. However, PKM retrieves memory vectors instead of functions, thus their values cannot vary according to the inputs. As we show in Section \ref{sec:experiment}, by changing the memory vectors to input-dependent expert networks, PEER can achieve significantly higher efficiency than PKM. Finally, \citet{csordas2023approximating} presented a unified view encompassing FFW, MoE and PKM and proposed to change the router normalization function in MoE and PKM from softmax to sigmoid or ReLU. 

\section{Conclusion}
This work introduces a fine-grained MoE architecture that decomposes an extremely wide dense feedforward layer into a large number of small experts. This design is supported by the recent discovery of the fine-grained MoE scaling law. To overcome the computational overhead of routing to a large number of experts, we apply the product keys to efficiently select a small subset of hidden neurons within a wide MLP layer. Empirical analysis using language modeling tasks demonstrate that given the same compute budget, PEER significantly outperforms dense transformers, coarse-grained MoEs and product key memory layers.

\section*{Acknowledgments}
The author would like to thank Adam Santoro, Arthur Guez, Arthur Szlam, Andrei Rusu, Marc'aurelio Ranzato, Simon Schug, Utku Evci, Doina Precup and Razvan Pascanu for their insightful discussions and invaluable advice. The author is also grateful to Zhitao Gong, Daniel Toyama, Qixuan Feng and Jiajun Shen for their technical assistance. Special thanks are due to Adam Santoro for sharing the isoFLOP analysis scripts and to Andy Brock for building and maintaining the internal codebase used to train the models.

\bibliography{main}
\bibliographystyle{tmlr}

\end{document}